\documentclass{article}
\usepackage{spconf}
\usepackage{amsmath,graphicx}
\usepackage{amssymb}
\usepackage{bbm}
\usepackage{textcomp}
\usepackage{mathtools}
\usepackage{booktabs}
\usepackage[utf8]{inputenc}
\usepackage[dvipsnames]{xcolor}
\usepackage{mystyle}
\usepackage{multirow}
\usepackage{tikz}
\usepackage{hyperref}
\usepackage{pgfplots}
\usepackage{tikz-3dplot}
\usetikzlibrary{arrows.meta}
\usetikzlibrary{positioning,through, backgrounds,decorations.pathreplacing,shapes.misc,patterns}
\pgfplotsset{compat=1.17}
 
\tdplotsetmaincoords{70}{30}

\title{USTED: Improving ASR with a Unified Speech and Text Encoder-Decoder}
\name{Bolaji Yusuf$^{1,2}$\sthanks{Work done as an applied scientist intern at Amazon Alexa}, Ankur Gandhe$^1$, Alex Sokolov$^1$}
\address{
$^{1}$ Amazon Alexa, USA \\
$^{2}$ Boğaziçi University, Department of Electrical and Electronics Engineering, Istanbul, Turkey \\
$^{3}$ Brno University of Technology, Faculty of Information Technology, Speech@FIT, Czechia}
\begin{document}
\ninept
\maketitle
\begin{abstract}
Improving end-to-end speech recognition by incorporating external text data has been a longstanding research topic. There has been a recent focus on training E2E ASR models that get the performance benefits of external text data without incurring the extra cost of evaluating an external language model at inference time. In this work, we propose training ASR model jointly with a set of text-to-text auxiliary tasks with which it shares a decoder and parts of the encoder. When we jointly train ASR and masked language model with the 960-hour Librispeech and Opensubtitles data respectively, we observe WER reductions of 16\% and 20\% on test-other and test-clean respectively over an ASR-only baseline without any extra cost at inference time, and reductions of 6\% and 8\% compared to a stronger MUTE-L baseline which trains the decoder with the same text data as our model. We achieve further improvements when we train masked language model on Librispeech data or when we use machine translation as the auxiliary task, without significantly sacrificing performance on the task itself.
\end{abstract}
\begin{keywords}
sequence-to-sequence, multitask, end-to-end ASR, masked language model, machine translation
\end{keywords}
\section{Introduction}
\label{sec:intro}
End-to-end (E2E) approaches to ASR arose as a more direct alternative to hybrid methods. By leveraging the considerable expressive power of large neural networks, E2E models obviate the need to train multiple disparate models while also considerably simplifying the onerous task of building an ASR decoder. This simplification, however, comes at a cost of data efficiency. Where the various modules of the hybrid approach can be estimated separately with various sources of data, doing the same for E2E models in a principled way is still an open problem.

Several works have attempted to improve E2E ASR by incorporating external data. Typically, this involves training with unpaired speech by using pseudo labels from an ASR system~\cite{vesely2013semi,kahn2020self,khurana2021unsupervised}, pretraining with surrogate unsupervised objectives~\cite{baevski2020wav2vec,Khurana2020Conv,liu2021tera} or combining these into an ASR-TTS cycle consistency objective~\cite{tjandra2017listening,baskar19_interspeech}.

Another line of work is concerned with incorporating unpaired text into the ASR model through fusion with an external language model (LM) ~\cite{gulcehre2015using,Chorowski2017} or hallucinating corresponding speech to increase the pool of available ASR training data~\cite{rossenbach2020generating,wang2020improving,baskar2021eat}. Recently, there has been a focus on manipulating the LM induced by the decoder of an E2E ASR model~\cite{variani2020hybrid}. In~\cite{mcdermott2019density,meng2021internal}, the decoder LM score is subtracted before shallow fusion with an external LM to improve domain adaptation. In~\cite{wang21t_interspeech} on the other hand, the decoder is trained with a large text corpus alleviating the need for an external LM.

Inspired by T5~\cite{raffel2020exploring} which uses a multitask generative construction, in this paper, we take the view that ASR is not a problem in a vacuum but one of a set of interrelated sequence transduction tasks. Based on the observation that various sequence-to-sequence tasks can be framed as trying to generate text from their respective input modalities (e.g. ASR is the task of generating text from speech, MT is the task of generating text from text in another language etc.), we will use a set of sequence-to-sequence tasks with the same output language (English in our case) to train a single model which will generate the correct text irrespective of the modality of its input.

We propose the Unified Speech and Text Encoder-Decoder (USTED), an attention-based model trained on ASR along with a variety of text-to-text transduction tasks. Instead of a single speech encoder, USTED has a bank of shallow task-specific \textit{modality} encoders which transform their respective inputs into a shared space from which a task-agnostic \textit{context} encoder operates on them. Finally, a shared decoder autoregressively generates the desired text. The crux of our approach is sharing the parameters of the context encoder and the decoder across all tasks, a choice which we hypothesize will improve the ASR in two ways: we'll get the advantages of training the decoder on a larger text corpus since the text tasks contain a much larger number of training sentences than the ASR; similar to models pretrained on unlabeled speech, we'll get a better encoder by training the parameters on a surrogate objective (text-to-text). Moreover, from a wider perspective, we get a model capable of solving multiple tasks simultaneously, and benefits from doing so, without significantly increasing the number of parameters.

We conduct experiments on Librispeech  and Opensubtitles which show that the proposed method offers an effective way of incorporating external text for improving ASR without requiring an external language model. Furthermore, we show masked language modeling to be a suitable text-to-text task for our purposes, thereby eliminating the need to search for text-to-text tasks with paired data.

\section{Model}
\label{sec:model}
\subsection{Attention-based encoder-decoder for ASR}
The model proposed in this paper is based on the attention encoder-decoder (AED) paradigm for end-to-end ASR~\cite{chan2016listen,bahdanau2016end}. An AED model with encoder parameters $\Matrix{\phi}$ and decoder parameters $\Matrix{\theta}$ directly models the distribution:
\begin{align}
    p_{\Matrix{\xi}}(\Matrix{y}|\Matrix{X}) = \prod_{u=1}^{U_y} p_{\Matrix{\xi}}(y_u|\Matrix{X}, y_{1:u-1}),
\end{align}
where $\Matrix{\xi} = \{\Matrix{\phi}, \Matrix{\theta}\}$, $\Matrix{X} := (\Matrix{x}_1, \dots, \Matrix{x}_N) \in \mathbb{R}^{N \times D_x}$ is a sequence of acoustic features and $\Matrix{y} := (y_1, \dots, y_{U_{{y}}})$ is the sequence of corresponding transcripts in the form of words, characters, or character-based subword units.

The model's encoder transforms $\Matrix{X}$ into a hidden representation $\Matrix{H} :=  (\Matrix{h}_1, \dots, \Matrix{h}_N) \in \mathbb{R}^{N \times D_h} = \Matrix{\phi}(\Matrix{X})$. The decoder recursively models the distribution $p_{\Matrix{\theta}}(y_u|\Matrix{X}, {y}_{1:u-1})$
of the $u$-th token conditioned on the encoder output $\Matrix{H}$ and previously generated tokens ${y}_{1:u-1}$. At each decoding step, the encoder outputs are summarized into a context vector $\Matrix{c}^{\Matrix{\phi}}_u(\Matrix{X}) = \sum_{n=1}^N w_{un} \Matrix{h}_n$, where the weights $\{w_{un}\}$ are computed with an additive attention mechanism~\cite{bahdanau2016end}. The context vector is fed into the decoder along with a trainable embedding of the output at the previous step so that the conditional distribution is simplified to:
\begin{align}
    p_{\Matrix{\xi}}(\Matrix{y}|\Matrix{X}) = \prod_{u=1}^{U_y} p_{\Matrix{\theta}}(y_u|\Matrix{c}_u^{\Matrix{\phi}}(\Matrix{X}), {y}_{u-1}).
\end{align}

The model is trained with the cross-entropy objective which minimizes the negative log-likelihood of the correct transcriptions $\hat{\Matrix{{y}}}$ of data $\Matrix{X}$ in a training set $\mathcal{D}$:
\begin{align}
    \mathcal{L}_{asr} = -\sum_{\Matrix{X}, \hat{\Matrix{{y}}} \in \mathcal{D}}\sum_{u=1}^{U_{\hat{y}}} \log p_{\Matrix{\theta}}(\hat{y}_u|\Matrix{c}_u^{\Matrix{\phi}}, \hat{{y}}_{u-1}),
    \label{eqn:loss_asr}
\end{align}
where the explicit dependence of $\Matrix{c}_u^{\Matrix{\phi}}(\cdot)$ on $\Matrix{X}$ is henceforth dropped from the notation to reduce clutter.

\begin{figure}
    \centering
    \includegraphics[width=0.95\linewidth]{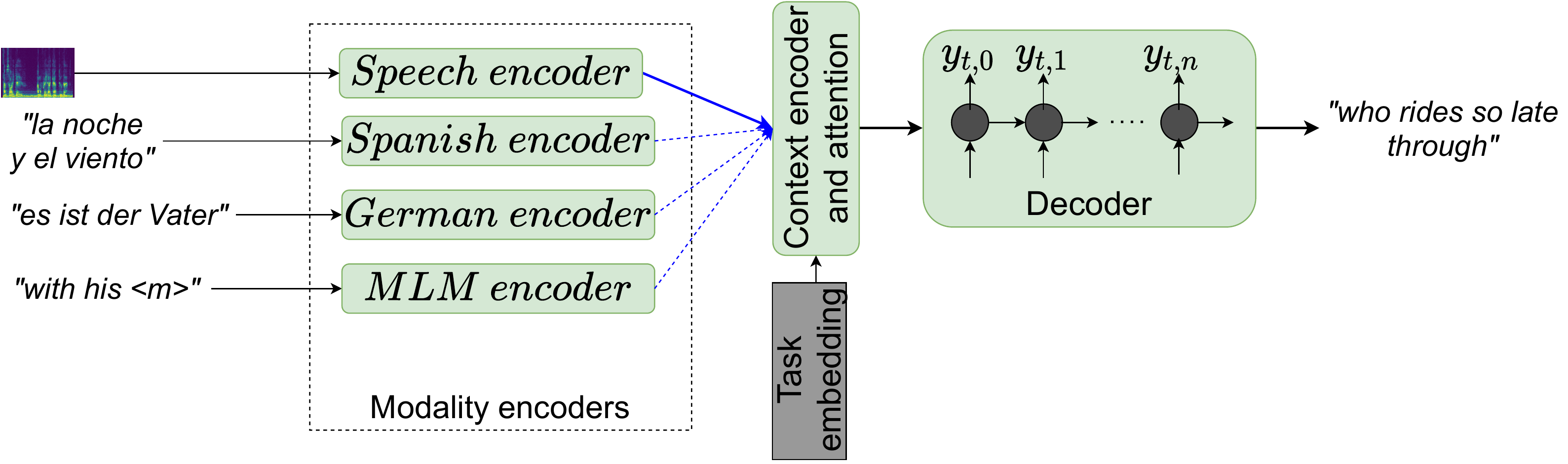}
    \caption{Schema of the proposed approach. A bank of shallow modality encoder project an input feature into a space from which a shared encoder and decoder generate the corresponding textual output.}
    \label{fig:model}
\end{figure}

\subsection{Unified speech and text encoder-decoder}
USTED, depicted in Figure~\ref{fig:model}, is a modified AED architecture which simultaneously operates on speech as well as a variety of text-to-text transduction tasks. Given a set of transduction tasks $\{T^1, \dots, T^Q\}$ with associated data:
\begin{align*}
    \mathcal{D} = \Bigl\{\mathcal{D}^1 = \{\Matrix{X}, \Matrix{y}\}^1, \mathcal{D}^2 = \{\Matrix{X}, \Matrix{y}\}^2, \dots, \mathcal{D}^Q = \{\Matrix{X}, \Matrix{y}\}^Q\Bigr\},
\end{align*}
The model is composed of a bank of task-specific \textit{modality} encoders $\{\Matrix{\phi}^1, \dots, \Matrix{\phi}^Q\}$, a \textit{context} encoder $\Matrix{\phi}$ which is shared across tasks, and a single, task-agnostic decoder $\Matrix{\theta}$.

For a data sample $\Matrix{X}$ from the $q$-th task, the corresponding modality encoder transforms the input into a shared space from where it is further transformed by the shared encoder:
 \begin{align}
     \tilde{\Matrix{H}}:= (\tilde{\Matrix{h}}_1, \dots, \tilde{\Matrix{h}}_N) &= \Matrix{\phi}^q(\Matrix{X}) \nonumber \\
     \Matrix{H} &= \Matrix{\phi}(\tilde{\Matrix{H}}).
     \label{eqn:task_specific_encoder}
 \end{align}
 Note that for tasks with textual input, a trainable, task-specific, embedding layer is used to embed the discrete input symbols into a continuous space.
Finally, as in regular AED, the decoder autoregressively generates textual symbols with context vectors $\Matrix{c}^{\Matrix{\phi}, \Matrix{\phi}^q}_u(\Matrix{X}) = \sum_{n=1}^N w_{un} \Matrix{h}_n$ computed from the encoder outputs.

\subsubsection{Training method and objective}
We train USTED with the same teacher-forced cross-entropy objective as the regular AED with the caveat that now we compute the loss for data across different tasks. Thus, we minimize the negative log-likelihood of generating the correct output sequences:
\begin{align}
    \mathcal{L}_{gen} = -\sum_{q=1}^{Q}\sum_{\Matrix{X}, \hat{\Matrix{{y}}} \in \mathcal{D}^q}\sum_{u=1}^{U_{\hat{y}}} \log p_{\Matrix{\theta}}(\hat{y}_u|\Matrix{c}_u^{\Matrix{\phi}, \Matrix{\phi}^q}, \hat{{y}}_{u-1}).
    \label{eqn:loss_gen}
\end{align}

When training, we uniformly sample from the set of tasks and then sample data within the task. Thus, each batch only has data from a single task.

\subsubsection{Task embedding}
\label{sec:task_embedding}
We experiment with cluing in the decoder (and shared encoder) on what task the current sample belongs to. We achieve this by prepending a trainable task-specific embedding, $\tilde{\Matrix{h}}_0^q$, to the output of the task specific encoder modifying~\eqref{eqn:task_specific_encoder} to:
\begin{align}
    \tilde{\Matrix{H}}:= \Bigl(\tilde{\Matrix{h}}_0^q, \tilde{\Matrix{h}}_1, \dots, \tilde{\Matrix{h}}_N\Bigr) = \Bigl(\tilde{\Matrix{h}}_0^q, \Matrix{\phi}^q(\Matrix{X})\Bigr).
\end{align}
We hypothesize that this would allow us to better generate text from the output distributions of the various tasks and we use it in all our experiments except where otherwise specified.



\section{Experiments}
\label{sec:experiments}
\begin{table}[b]
    \centering
    \caption{Trigram perplexities of Librispeech and Opensubtitles language models evaluated on Librispeech and Opensubtitles dev sets.}
    \begin{tabular}{lccc}
    \toprule
         LM training data & dev-clean & dev-other & test-es-\emph{en} \\
         \midrule
         Librispeech-960h & 264 & 230 & 409 \\
         Librispeech-40m & 175 & 163 & 329 \\
         Opensubtitles-es-\emph{en} & 476 & 426 & 131 \\
         \bottomrule
    \end{tabular}
    \label{tab:ppl}
\end{table}

\begin{table*}[t]
    \caption{WER comparison between USTED and various baselines. Where applicable, the parenthesized R and K values denote the masking rate and the number of shared encoder layers respectively.}
    \centering
    \begin{tabular}{lccccccc}
    \toprule
         Model & Auxiliary dataset & Auxiliary task & dev-clean & dev-other & test-clean & test-other & os-en-tts \\
         \midrule
         Baseline & - & - & 5.6 & 14.2 & 5.9 & 15.2 & 18.1 \\
         MUTE-L~\cite{wang21t_interspeech} & Opensubtitles & - & 4.9 & 12.7 & 5.1 & 13.5 & 14.7 \\
         USTED (K=3) & Opensubtitles & de-en MT & 4.5 & 12.1 & 4.6 & 12.9 & 14.3 \\
         USTED (K=1) & Opensubtitles & \textbf{es-en MT} & \textbf{4.4} & \textbf{12.0} & \textbf{4.5} & \textbf{12.5} & \textbf{14.3} \\
         USTED (K=1, R=0.4) & Opensubtitles & MLM & 4.6 & 12.1 & 4.7 & 12.7 & 14.3 \\
         USTED (K=3, R=0.4) & Opensubtitles & MLM + MT & 4.6 & 12.1 & 4.7 & 12.6 & 14.0 \\
         \midrule
         MUTE-L & Librispeech-40m & - & 4.6 & 12.3 & 4.9 & 12.9 & 15.3 \\
         USTED (K=3, R=0.4) & Librispeech-40m & MLM & 4.5 & 11.9 & 4.7 & 12.5 & 14.9 \\
         \bottomrule
    \end{tabular}
    \label{tab:test_performance}
\end{table*}

\subsection{Experiment setup}
\subsubsection{Tasks and datasets}
\textbf{Speech recognition:} We use the full 960 hours data from the Librispeech corpus~\cite{panayatov2015librispeech} for training the ASR task. We use the various test splits from the same corpus for evaluation. In particular, we use the sum of word error rates (WER) on \texttt{dev-other} and \texttt{dev-clean} as the criterion for selecting best training checkpoints for the \texttt{test-clean}, \texttt{test-other} splits. We also evaluate on the \texttt{os-en-tts} set obtained by synthesizing speech from the 100k sentences of the English part of \texttt{test-es-en} described below with the TTS system described in~\cite{fazel21_interspeech}.

\noindent\textbf{Machine translation:} As one of the side tasks, we explore MT, for which we take the paired Spanish-English (\texttt{es-en}) and German-English (\texttt{de-en}) sets from the Opensubtitles corpus~\cite{lison-tiedemann-2016-opensubtitles2016} of aligned movie subtitles amounting to 61 million and 22 million sentences respectively. From each of these, we create splits of 100k sentences each (\texttt{test-es-en}, \texttt{test-de-en}) for evaluating MT performance, with the remaining data used for training. The training and evaluation splits are constructed such that no two splits contain sentences from the same movie. Note that, as indicated in Table~\ref{tab:ppl}, the contemporary, conversational language in the Opensubtitles data differs significantly from the literary, often archaic, forms in Librispeech, reflected in the perplexity gaps between using the Librispeech LM on Opensubtitles dev set and vice versa.

\noindent \textbf{Masked language modeling}: The final training task that we explore is masked language modeling where we use the English part (61 million sentences) of the Spanish-English Opensubtitles. In this task, we corrupt a sentence by randomly masking some words and then requiring the model to reconstruct the \textit{uncorrupted} sentence. Note that we reconstruct the entire sequence rather than just the masked tokens since the decoder is shared with other auto-regressive generative tasks and thus should be trained as a proper language model.

\subsubsection{Model configuration}
\label{sec:config}
Our implementation of USTED is based on the listen, attend and spell architecture~\cite{chan2016listen} with a 4-layer BiLSTM encoder and a two-layer LSTM decoder with four-headed additive attention. Each encoder layer has 1024 units in each direction.
As described above, the decoder is shared across all tasks as is part of the encoder. Note that the 4 encoder layers include both the task-specific modality encoders ($K$ layers) and the shared context encoder ($4-K$ layers). For instance, when we share 3 layers, this means the shared context encoder has 3 BiLSTM layers and the modality encoders have 1 BiLSTM layers each. Thus, while the total number of parameters in the model reduces as the number of shared layers is increased, each task always has the same number of parameters available to it regardless of the number of shared layers.

The network outputs tokens from a unigram subword tokenizer~\cite{kudo2018subword} with {\NumTokenizers} tokens trained on Librispeech. For English-to-English text tasks, we use the same tokenizer for the input. For the translation tasks, we train a tokenizer with {\NumTokenizers} for the input language. The input text tokens are are embedded by task-specific embedding layers into 192 dimensional space. The input to the speech modality encoder are 64-dimensional log-filterbank features stacked with the two frames to the left and then downsasmpled by 3 resulting in a 192-dimensional input feature for every $30ms$.

\subsubsection{Speech encoder pretraining}
In preliminary experiments, we found that getting competitive ASR results required oversampling the ASR task. This however degraded performance on other tasks. We solve this issue by pretraining a standalone LAS ASR system whose encoder's parameters are transferred to initialize both the speech modality encoder 
and the context encoder. 
We randomly initialize the decoder and the other tasks' modality encoders as we did not observe any benefits in pretraining them. After pretraining, we sample training tasks uniformly, leading to faster convergence and better performance on auxiliary tasks.


\subsection{Test set performance}
Table~\ref{tab:test_performance} shows a comparison of USTED performance with two baselines: a standalone LAS model with no auxiliary tasks and MUTE-L which uses the text data to further train the decoder as described in ~\cite{wang21t_interspeech}. All models use the Librispeech 960h for ASR training. We consider Opensubtitles and Librispeech as the source of auxiliary text data for both MUTE-L and USTED. Note that in the Opensubtitles setting, we use the English part of the Spanish-English translation corpus for training MUTE-L as well as MLM for USTED.

When we use Opensubtitles as the auxiliary dataset, we observe that all the variants of USTED improve upon MUTE-L which already outperforms the ASR-only baseline by a significant margin. When we use Spanish-English translation as the auxiliary task, we observe relative word error rate improvements of {\EsEnTestClean\%} and {\EsEnTestOther\%} on \texttt{test-clean} and \texttt{test-other} respectively. While using MLM as the auxiliary task for USTED results in more modest improvements, it has the advantage of only requiring the same unpaired text data as MUTE-L unlike machine translation task requiring paired data that is normally harder to obtain. Therefore, we'll conduct most of our further analyses on the MLM task.

Using Librispeech as the auxiliary dataset precludes us from using machine translation task since we have no paired data. USTED with MLM still outperforms MUTE-L in all test sets in this setting. We note that when compared to the corresponding USTED models trained with Opensubtitles auxiliary data, both models perform better on Librispeech test sets and worse on the Opensubtitles test sets; this supports the intuition that having a task from domain matching the test set leads to better performance.

Finally, we note that using Spanish-English translation as the auxiliary task achieves comparable performance with MLM and outperforms MUTE-L on Librispeech test sets even when the latter are trained with Librispeech text, while being significantly better than both on the synthesized Opensubtitles test set. This indicates that the performance gains realized from using a suitable task can make up for differences in output domain. However, finding such a suitable task \emph{a priori} for any target test set remains an open question.

\begin{table}[b]
    \caption{WER of USTED as we vary the number of encoder layers that are shared by ASR and MLM.}
    \centering
    \begin{tabular}{lccccc}
    \toprule
         Shared layers & 0 & 1 & 2 & 3 & 4 \\
         \midrule
         dev-clean & 5.0 & 4.7 & 4.9 & 4.7 & 4.8\\
         dev-other & 12.8 & 12.2 & 12.1 & 12.5 & 12.7 \\
         \bottomrule
    \end{tabular}
    \label{tab:shared_layers}
\end{table}

\begin{table}[t]
    \caption{WER as the MLM masking rate is varied.}
    \centering
    \begin{tabular}{lcccccc}
    \toprule
         Masking ratio & 0 & 0.2 & 0.4 & 0.6 & 0.8 & 1\\
         \midrule
         dev-clean & 4.9 & 4.7 & 4.7 & 4.5 & 4.6 & 4.6 \\
         dev-other & 12.5 & 12.5 & 12.2 & 12.2 & 12.3 & 12.2 \\
         \bottomrule
    \end{tabular}
    \label{tab:masking_ratio}
\end{table}
\subsection{MLM performance analysis.}

\begin{figure}[t!]
    \centering
    \resizebox{0.29\textwidth}{!}{%
        \begin{tikzpicture}
            \input{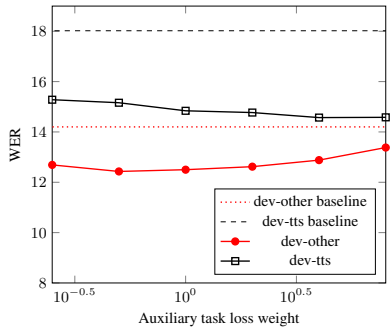}
        \end{tikzpicture}
    }
    \caption{WER on the \texttt{os-es-tts} set as the weight of the MLM training loss is varied.}
    \label{fig:loss_weight}
\end{figure}
As described in Section~\ref{sec:model}, in addition to sharing the decoder across all tasks, we also share the last few encoder layers. Table~\ref{tab:shared_layers} shows the impact of changing varying the number of shared encoder layers from 0 (only share the decoder) to 4 (sharing the \textit{entire} model) with the masking rate fixed to 0.4. We observe that sharing 0 layers performs worst with performance comparable to MUTE-L. The performance peaks with one shared encoder layer and starts to degrade as more layers are shared. This indicates that the gains obtained from shared information across tasks eventually get offset by the loss of capacity from reduction in total number of parameters.

Table~\ref{tab:masking_ratio} shows the effect of modifying the masking rate while keeping the number of shared encoder layers fixed to 1. The worst ASR performance is obtained when none of the input words are masked, i.e. having a sequence-to-sequence autoencoder as the auxiliary task. Performance improves with increasing masking rate with peak performance when 60\% of the input text is masked with drop in performance afterwards. Surprisingly, even a masking rate of 1 outperforms the MUTE-L baseline despite having using the same information plus the length of the input. We hypothesize that even with all tokens masked in the input, sharing the encoder regularizes it and we will explore this further in future work

In Equation~\ref{eqn:loss_gen}, each task is weighted equally and that is what we have used so far. Figure~\ref{fig:loss_weight} shows the impact on ASR of re-weighting the losses of the MLM task while keeping the ASR loss weight at 1. On \texttt{dev-other}, we observe that the WER is best at 0.5 and starts to worsen as the weight is increased. We observed a similar trend on \texttt{dev-clean}, although we elected to exclude it from the figure to avoid warping the scale of the graph. On \texttt{dev-tts}, which is 24\% of the entire \texttt{os-en-tts} set and is thus from the same domain as the MLM text, we observe that the WER improves as we increase the MLM weight up to a weight of 8 which is the maximum we tried.
\begin{table}[t]
    \caption{Impact of task embedding on ASR dev-set performance.}
    \centering
    \begin{tabular}{lcccc}
    \toprule
          & \multicolumn{2}{c}{MLM} & \multicolumn{2}{c}{MLM + MT} \\
          & clean & other & clean & other\\
         \midrule
         USTED & 4.8 & 12.2 & 4.6 & 12.1 \\
         \ \ - task embedding & 4.8 & 12.3 & 4.7 & 12.2 \\
         \bottomrule
    \end{tabular}
    \label{tab:emb_adv}
\end{table}

\begin{table}[t]
    \caption{BLEU scores achieved on Spanish and German machine translation tasks.}
    \centering
    \begin{tabular}{lcccc}
    \toprule
          & test-es-en & test-de-en \\
         \midrule
         Baseline & 31.6 & 26.3 \\
         USTED (MT) & 31.5 & 25.8 \\
         USTED (MLM + MT) & 29.6 & 24.2 \\
         \ \ - task embedding & 23.1 & 18.5 \\
         \bottomrule
    \end{tabular}
    \label{tab:emb_mt}
\end{table}

\subsection{Effect of task embedding}
In all our experiments so far, we have prepended a task embedding to the output of each modality encoder. Tables~\ref{tab:emb_adv}~and~\ref{tab:emb_mt} show the impact of the task embeddings on ASR and translation performance. While the impact on ASR performance is minimal, the effect on MT is more drastic. Using the task embedding allows us to improve ASR without sacrificing much in terms of MT performance when compared to a baseline model trained on MT alone. Note that the performance of USTED on MT is on par with the standalone baseline when it is trained with two tasks, i.e. ASR and the relevant language pair.

\subsection{ASR error qualitative analysis}
Table~\ref{tab:qualitative} contains examples of some errors corrected by USTED. We observe that USTED reduces instances of acoustically plausible but linguistically nonsensical transcriptions, although we found that MUTE-L also makes similar corrections. While USTED fixes other kinds of errors, we could not find any conclusive patterns.
\begin{table}[t!]
\caption{Examples of errors corrected by USTED. Errors made by the baseline are in red and corrections made by USTED are in blue.}
    \centering
    \begin{tabular}{ll}
    \toprule
    System & Transcription \\
    \midrule
         Baseline &  THEY SAY ILLUMINATION BY \\ & \textcolor{red}{CANNOLIDAYS} THE PRETTIEST IN THE WORLD \\
         USTED & THEY SAY ILLUMINATION BY \textcolor{blue}{CANDLE LIGHT} \\ & 
         \textcolor{blue}{IS} THE PRETTIEST IN THE WORLD \\
         \midrule
         Baseline & DON'T WORRY \textcolor{red}{SAYS ODEIR} IT'LL ALL COME \\ & RIGHT PRETTY SOON \\
         USTED & DON'T WORRY \textcolor{blue}{SIZZLE DEAR} IT'LL ALL COME \\ & RIGHT PRETTY SOON \\
         \bottomrule
    \end{tabular}
    \label{tab:qualitative}
\end{table}

\section{Conclusions}
\label{sec:conclusions}
In this work, we've introduced a framework for jointly training ASR with text-to-text transduction tasks wherein parameters of an attention-based sequence to sequence model are shared across tasks. We have shown the efficacy of using machine translation and masked language modeling as the auxiliary tasks. Our model achieves significant improvements in ASR performance without incurring any computational overhead at inference time. Moreover, the added training cost is further justified by the fact that our multitask model also performs the auxiliary tasks, specifically machine translation, with minimal degradation in performance. We leave further exploration and analysis of auxiliary tasks from other domains and modalities as subject of future work.

\bibliographystyle{IEEEbib}
\bibliography{refs}

\end{document}